%% file: root.tex
\documentclass[letterpaper, 10 pt, conference]{ieeeconf}  

\input{preamble}

\IEEEoverridecommandlockouts                              

\title{\LARGE \bf
Initialization of Monocular Visual Navigation for Autonomous Agents Using Modified Structure from Small Motion
}

\author{
Juan-Diego Florez$^{1}$, Mehregan Dor$^{1}$, and Panagiotis Tsiotras$^{1}$
\thanks{This work has been supported by Verus Research
under AFRL contract No. FA9453- 23-C-A025 and
by NSF award FRR-2101250.}
\thanks{Any opinions, findings and conclusions or recommendations expressed in this material are those of the author(s) and do not necessarily reflect the views of the Air Force Research Laboratory.} %
\thanks{$^{1}$School of Aerospace Engineering, Georgia Institute of Technology, Atlanta, GA 30332, USA {\tt\small \{jdflorez, mehregan.dor, tsiotras\}@gatech.edu}}
}

\renewcommand{\baselinestretch}{0.98} 

\begin{document}

\normalsize

\maketitle
\thispagestyle{empty}
\pagestyle{empty}

\begin{abstract}

We propose a standalone monocular visual Simultaneous Localization and Mapping (vSLAM) initialization pipeline for autonomous space robots. Our method, a state-of-the-art factor graph optimization pipeline, extends Structure from Small Motion (SfSM) to robustly initialize a monocular agent in spacecraft inspection trajectories, addressing visual estimation challenges such as weak-perspective projection and center-pointing motion, which exacerbates the bas-relief ambiguity, dominant planar geometry, which causes motion estimation degeneracies in classical Structure from Motion, and dynamic illumination conditions, which reduce the survivability of visual information. We validate our approach on realistic, simulated satellite inspection image sequences with a tumbling spacecraft and demonstrate the method's effectiveness over existing monocular initialization procedures.

\end{abstract}

\section{INTRODUCTION}

\subsection{Problem Statement}\label{sec:problemstatement}
Accurate estimation of the relative pose and 3D map of a non-cooperative resident space object (RSO) enables the real-time guidance and control required for missions such as satellite repair and active debris removal and is crucial for safe inspection and proximity operations~\cite{Ventura2017, Ma2023, Albee2021, astroscaleADRASJAstroscale, Aglietti2019}. 
This work addresses the initialization of a relative navigation pipeline on an autonomous chaser spacecraft tracking a non-cooperative RSO, without prior knowledge of the RSO's kinematics, dynamics, or 3D structure.

Monocular visual Simultaneous Localization and Mapping (vSLAM) systems provide real-time estimates, remain robust in dynamic environments, and operate with low power consumption and mass. 
Unlike other approaches, they do not require ranging (LIDAR), time-of-flight detection or structured light projection (RGB-D cameras), or specialized calibration (stereo cameras). 
Although monocular vSLAM systems are prone to scale and depth ambiguities, proper initialization and use of advanced estimation algorithms enable accurate relative motion, 3D mapping, and dynamic motion characterization~\cite{Dor2018, Dor2024}.

Initializing RSO inspection trajectories is challenging due to the weak-perspective projection caused by large operating ranges~\cite{hartley2003multiple}, which limits depth variation and exacerbates the bas-relief ambiguity.
Under these conditions, small rotations, small translations, and depth scaling yield similar 2D projections, thereby complicating 3D mapping. 
The ambiguity is further pronounced in center-pointing motions, where reduced apparent lateral motion further complicates the disambiguation of motion components.

Both feature-based and pixel-based detection approaches face additional challenges.
For instance, dynamic illumination conditions in space limit the survivability of tracked features and reduce the reliability of photometric error in pixel-based approaches.
RSOs often have dominant planar geometries (e.g., solar panel arrays), leading to degeneracies in fundamental matrix estimation~\cite{chum2005two}, and low-texture areas that reduce pixel-intensity variation and hinder pixel-based methods.

To address these challenges, we enhance the Structure from Small Motion (SfSM)~\cite{Yu2014-ni, Ha2018-et} framework to develop a monocular vSLAM initialization module that is robust to the ambiguities and degeneracies of inspection trajectories, enabling rapid convergence to accurate relative poses estimates and a high quality 3D mapping solution.

\subsection{Related Work}\label{sec:relatedwork}

Sparse, feature-based visual estimation methods are well-suited to online space-bound applications due to their computational efficiency over dense methods and their robustness to dynamic lighting conditions.
Assuming a known calibrated camera intrinsic matrix, the classical feature-based initialization, often found in Structure from Motion (SfM) schemes, involves estimating motion from two-view geometry~\cite{Davison2007}.
The 5-point algorithm~\cite{nister2004efficient}, commonly used in SLAM initialization, estimates the essential matrix for relative pose estimation and is often paired with Random Sample Consensus (RANSAC) for outlier rejection.
The essential matrix is decomposed into four possible relative pose solutions, refined through 3D point triangulation and cheirality checks.  
However, dominant planes in the scene can cause degeneracies in fundamental and essential matrix estimation, resulting in failed pose recovery and mapping errors. 
In such cases, homography-based methods~\cite{hartley2003multiple} are preferred. 

Model selection strategies can enhance the robustness of visual estimation pipelines. For example, ORB-SLAM~\cite{mur2015orb} solves for both fundamental matrix and homography transformation during initialization, using an error metric to select the best-fitting model. 
Similarly, the USAC\_FM\_8PTS algorithm~\cite{Raguram2013, opencv_library} applies DEGENSAC~\cite{chum2005two, Jin2020} to detect planar degeneracies and select the most appropriate model, and uses LOSAC~\cite{lebeda2012fixing} to refine the model through local optimization.
However, defining a unified metric to compare different models can yield inconsistent results~\cite{Torr97}, as 
higher degree-of-freedom (DoF) models may produce lower errors than simpler models without necessarily providing accurate motion recovery, particularly in noisy conditions.

Model-based methods require sufficient parallax between queried frames to generate accurate estimates.
Consequently, pipelines using this approach delay initialization until enough parallax is present to resolve depth, necessitating a wide baseline for estimation; hence, they are known as ``delayed initialization" and "wide-baseline" methods. 
However, at operating ranges with weak-perspective projection, achieving sufficient parallax requires significant relative motion between successive camera frames.
Accordingly, delayed initialization faces a timing dilemma---delaying too long risks losing feature tracks due to changing illumination conditions over a large baseline motion, while initializing early without sufficient parallax leads to unreliable estimates. Thus, wide-baseline approaches are not amenable to small-motion image sequences~\cite{DorThesis}.

Sensor-fusion approaches, like visual-inertial SLAM~\cite{Qin2018, Campos2021-vp}, may mitigate some limitations of purely visual methods. However, inertial measurements are ineffective for estimating motion in free-fall---typical of un-powered inspection orbits~\cite{Dor2024}---and therefore do not provide useful cues for pose initialization in RSO inspection trajectories.

Structure from Small Motion (SfSM)~\cite{Yu2014-ni} offers a non-delayed, purely visual approach that leverages small apparent motion between consecutive frames. 
Rather than waiting for large baselines to accumulate, SfSM utilizes the available visual information early, thereby reducing the risk of losing track of critical features to changing illumination conditions and motion. 
Thus, SfSM is especially effective for early initialization, where the robust feature tracking is crucial. 
SfSM is well-suited to handle small-motion scenarios, avoiding the dependence on wide-baseline parallax and enabling faster, more reliable initialization in spacecraft inspection arcs.

To improve the robustness of SfSM under the small-motion, small-angle rotation assumption, Ha et al.~\cite{Ha2018-et} propose a three-step estimation pipeline:
first, camera rotations are recovered using RANSAC; second, camera translations and 3D point inverse depths are estimated through a restricted bundle adjustment (BA); and finally, a full BA is performed. 
Our work extends this three-step SfSM framework to robustly initialize under the challenging visual estimation conditions of RSO inspection trajectories.
 
\subsection{Contributions}

In this work, we develop a three-step SfSM monocular initialization pipeline that modifies the small-motion assumption to achieve a robust and accurate initial map and relative trajectory solution for space vSLAM applications.
We extend the approach in~\cite{Ha2018-et} to ensure robust initialization in RSO inspection trajectories, which are challenged by weak-perspective projection, center-pointing motion, dominant planar geometry, and dynamic illumination.
We validate the performance of our initialization method using realistic simulated image sets.

Our specific contributions to the SfSM procedure are as follows:
\begin{itemize}
    \item We redefine the small-motion, small-angle rotation assumptions of previous SfSM pipelines to address weak-perspective projection and center-pointing motion, where the apparent motion of small translations is comparable to that of small rotations.
    \item We re-parameterize inverse depth to guarantee the cheirality condition and ensure numerical stability.
    \item We reformulate 3D landmarks to account for image feature quantization by allowing in-camera-plane variation of their coordinates.
\end{itemize}

\section{Methodology}

The proposed initialization pipeline takes as input a SLAM front-end output consisting of feature-tracks that contain the pixel coordinates of \(m\) point features tracked across \(n\) frames, and processes this data in a three-step pipeline comprising:
\begin{enumerate}
  \item Rotation and scaled translation estimation
  \item Translation and inverse depth estimation
  \item Bundle adjustment
\end{enumerate}

\subsection{Notation}

For convenience, we set the 0-th camera frame as the reference frame and align it with the world frame. 
The camera motion from the reference frame to the \(i\)-th frame is described by a rigid transformation consisting of a rotation matrix \(\vec{R}_i\in \mathrm{SO}(3)\) and a translation vector \(\vec{r}_i\in \mathbb{R}^3\).
The coordinates of the $j$-th 3D landmark expressed in the $i$-th camera frame are denoted as $\vec{y}_{ij}$, and follow the relation
\begin{equation}
  \vec{y}_{ij} = \vec{R}_{i}\vec{y}_{0j}+\vec{r}_i,
  \label{eq:landmark1}
\end{equation}
where $\vec{y}_{0j} \triangleq [X_{j}, Y_{j}, Z_{j}]^\top$ is the 3D coordinate vector of the $j$-th landmark expressed in the reference frame (index~0).

For each \(j\)-th point and \(i\)-th image frame, we define the homogeneous pixel coordinates \(\vec{p}_{ij} = \left[u_{ij}, v_{ij}, 1\right]^\top\). 
We transform these coordinates to the camera coordinate frame using \(\vec{x}_{ij} = \vec{K}^{-1} \vec{p}_{ij}\), where \(\vec{K}\) is the pre-calibrated intrinsic matrix and \(\vec{x}_{ij} = \left[x_{ij}, y_{ij}, 1\right]^\top\).
The projected and normalized coordinate vector $\vec{x}_{ij}$ of the 3D landmark $\vec{y}_{ij}$ is given by $\vec{x}_{ij} =\langle \vec{y}_{ij} \rangle$, where $\langle \cdot \rangle$ denotes the normalization by the third component, such that $\langle [x,y,z]^\top \rangle = \left[ x/z, y/z,1 \right]^\top$ for $z\neq 0$. 
Measurement-related quantities are denoted using the superscript $(\cdot)^{\mathrm{m}}$. 

\subsection{Step 1: Rotation and Scaled Translation Estimation}\label{sec:step1}

Given $m$ sequences $(\vec{p}_{0j}^{\mathrm{m}},\vec{p}_{1j}^{\mathrm{m}}, \ldots, \vec{p}_{nj}^{\mathrm{m}}), \, j=1,\ldots,m$ of image point measurements $\vec{p}_{ij}^{\mathrm{m}}$ matched and tracked across image frames $i=1,\ldots,n$, we apply the RANSAC algorithm to obtain the best-fitting rotation estimate and corresponding inlier set between the reference frame and each of the frames $i=1,\ldots,n$.

As in~\cite{Ha2018-et}, we initially assume that the 3D coordinate of each landmark lies exactly along the back-projection of its corresponding image point, with some unknown depth value.
Consequently, $j$-th landmark, with expected coordinate vector \(\vec{y}_{0j}\), is parameterized by the camera frame coordinate measurement $\vec{x}_{0j}^{\mathrm{m}} = \vec{K}^{-1}\vec{p}_{0j}^{\mathrm{m}}$ and an estimated inverse depth $w_j$, such that \(\vec{y}_{0j} =\vec{x}_{0j}^{\mathrm{m}} / w_j\).
Thus, the inverse depth $w_j$ is the sole DoF determining the landmark's position.
Using this inverse depth parameterization, we rewrite~Eq.~\eqref{eq:landmark1}, yielding
\begin{equation}
  \vec{y}_{ij} = \vec{R}_{i}\frac{\vec{x}_{0j}^{\mathrm{m}}}{w_j}+\vec{r}_i.
  \label{eq:landmark2}
\end{equation}
Under the small motion assumption, we apply the first-order approximation \(\vec{R}_i\approx\mathit{I}_3 + \left[\vec{\theta}_i \right]_{\times}\)~\cite{Yu2014-ni}, where \(\mathit{I}_3\) is the \(3\times{3}\) identity matrix, \(\left[\cdot \right]_{\times}\) denotes the skew-symmetric matrix of a 3D vector, and \(\vec{\theta}_i = [\theta_{i1}, \theta_{i2}, \theta_{i3}]^\top\) and represents the rotation vector of the \(i\)-th camera frame. 
Hence,
\begin{align}
    \vec{R}_i = \begin{bmatrix}
        1 & -\theta_{i3} & \theta_{i2} \\
        \theta_{i3} & 1 & -\theta_{i1} \\
        -\theta_{i2} & \theta_{i1} & 1
    \end{bmatrix}.
\end{align}
From~Eq.~\eqref{eq:landmark2}, we obtain the expected normalized camera frame coordinates $\vec{x}_{ij} = \left[x_{ij}, y_{ij}, 1\right]^\top$, which correspond to the 3D point $\vec{y}_{ij}$.
Thus, $\vec{x}_{ij} = \langle \vec{y}_{ij} \rangle$, or, in scalar form,
\begin{equation}
\begin{split}
  x_{ij} &= \frac{x_{0j}^{\mathrm{m}} - \theta_{i3} y_{0j}^{\mathrm{m}} + \theta_{i2} + w_j r_{i1}}
  {-\theta_{i2} x_{0j}^{\mathrm{m}} + \theta_{i1} y_{0j}^{\mathrm{m}} + 1 + w_j r_{i3}}, \\
  y_{ij} &= \frac{\theta_{i3} x_{0j}^{\mathrm{m}} + y_{0j}^{\mathrm{m}} - \theta_{i1} + w_j r_{i2}}
  {-\theta_{i2} x_{0j}^{\mathrm{m}} + \theta_{i1} y_{0j}^{\mathrm{m}} + 1 + w_j r_{i3}}.
\label{eq:projected_point2}
\end{split}
\end{equation}
By virtue of normalization and essential to the derivation of ~Eq.~\eqref{eq:projected_point2}, we have $\langle \vec{R}_{i}\frac{\vec{x}_{0j}^{\mathrm{m}}}{w_j}+\vec{r}_i \rangle = \langle \vec{R}_{i}\vec{x}_{0j}^{\mathrm{m}}+w_j\vec{r}_i \rangle$.

Applying the small-translation assumption from~\cite{Ha2018-et}---where small rotations dominate apparent motion and scene points are distant---we assume $w_j\vec{r}_i\approx \vec{0}$, thereby simplifying ~Eq.~\eqref{eq:projected_point2}.
However, the center-pointing motion of RSO inspection trajectories implies that both translation and rotation can contribute similarly to the apparent motion of points in the image.
Consequently, \(\vec{R}_i \vec{x}_{0j} \sim w_j\vec{r}_i\), thereby invalidating the small-translation assumption. 

Accurately quantifying the rotation and translation contributions to the apparent motion is crucial for overcoming the bas-relief ambiguity. Hence, we apply weak-perspective projection and assume that, from the reference frame's perspective, the landmarks are tightly clustered at a large distance along the camera's boresight.  
We further assume the simplification $w_j \approx \bar{w}$ for each landmark $j$, and we re-parameterize 3D points such that
\begin{equation}
  \langle \vec{y}_{ij} \rangle = \langle \vec{R}_{i}\vec{x}_{0j}^{\mathrm{m}} + \Bar{\vec{r}}_{i} \rangle,
  \label{eq:landmark3}
\end{equation}
where \(\bar{\vec{r}}_{i} \triangleq \left[\bar{r}_{i1},\bar{r}_{i2}, \bar{r}_{i3}\right] =\Bar{w}\vec{r}_i\) represents the scaled translation.
Using this re-parameterization, the expected projected landmark coordinate $\vec{x}_{ij}$ has components:
\begin{equation}
    \begin{split}
          x_{ij} &= \frac{x_{0j}^{\mathrm{m}} - \theta_{i3} y_{0j}^{\mathrm{m}} + \theta_{i2} + \bar{r}_{i1}}
  {-\theta_{i2} x_{0j}^{\mathrm{m}} + \theta_{i1} y_{0j}^{\mathrm{m}} + 1 + \bar{r}_{i3}}, \\
  y_{ij} &= \frac{\theta_{i3} x_{0j}^{\mathrm{m}} + y_{0j}^{\mathrm{m}} - \theta_{i1} + \bar{r}_{i2}}
  {-\theta_{i2} x_{0j}^{\mathrm{m}} + \theta_{i1} y_{0j}^{\mathrm{m}} + 1 + \bar{r}_{i3}}.
\label{eq:projected_point3}
    \end{split}
\end{equation}

Following a RANSAC iteration and given a predetermined pixel distance threshold $\mu$, we select the best sample set rotation vector and scaled translation vector pair $(\vec{\theta}_i^\mathrm{s},\bar{\vec{r}}_i^\mathrm{s})$ that maximizes the cardinality of the inlier set $\mathcal{M}_i  \subset \{1,\ldots,m\}$, such that $\|\vec{p}_{ij}^{\mathrm{m}} - \vec{K}\vec{x}_{ij}\|<\mu$ for all $j\in \mathcal{M}_i$. 
Each step 1 optimizer pair $(\vec{\theta}_i^{*}, \bar{\vec{r}}_i^{*})$ minimizes the associated cost $\sum_{j\in \mathcal{M}_i} \|\vec{x}_{ij}^{\mathrm{m}} - \vec{x}_{ij}\|^2$, implying that $\vec{x}_{ij}^{\mathrm{m}} - \vec{x}_{ij}=\vec{0}$ for all $j\in \mathcal{M}_i$ by first-order optimality conditions. 
Substituting the expected point $\vec{x}_{ij}$ using~Eq.~\eqref{eq:projected_point3}, we obtain a system of equations linear in $\vec{\theta}_i$ and $\bar{\vec{r}}_i$, yielding
\begin{align}
    A_{ij}&
    \begin{bmatrix}
        \theta_{i1} & \theta_{i2} & \theta_{i3} & \bar{r}_{i1} & \bar{r}_{i2} & \bar{r}_{i3}
    \end{bmatrix}^\top
    = \begin{bmatrix}
        x_{0j}^{\mathrm{m}} - x_{ij}^{\mathrm{m}} \\ y_{0j}^{\mathrm{m}} - y_{ij}^{\mathrm{m}}
    \end{bmatrix},
    \label{eq:rotation3}
\end{align}
where
\begin{equation*}
    A_{ij} \triangleq 
    \begin{bmatrix}
        x_{ij}^{\mathrm{m}} y_{0j}^{\mathrm{m}} & -x_{ij}^{\mathrm{m}} x_{0j}^{\mathrm{m}} - 1 & y_{0j}^{\mathrm{m}} & -1 & 0 & x_{ij}^{\mathrm{m}} \\
        x_{ij}^{\mathrm{m}} y_{0j}^{\mathrm{m}} + 1 & -y_{ij} x_{0j}^{\mathrm{m}} & -x_{0j}^{\mathrm{m}} & 0 & -1 & y_{ij}^{\mathrm{m}}
    \end{bmatrix}. \nonumber 
\end{equation*}
The system in~Eq.~\eqref{eq:rotation3} can be solved in a least-squares sense using standard techniques such as matrix decomposition. 

The structure of the feature-track inputs to the RANSAC process ensures that all inliers between the reference frame and the \(n\)-th frame remain inliers across all intermediate frames. Thus, we perform RANSAC only between these two frames before solving~Eq.~\eqref{eq:rotation3} to estimate the relative rotations \(\vec{\theta}_i\) and scaled translations \(\Bar{\vec{r}}_i\) for each intermediate frame.
Our RANSAC procedure uses three points for robust triangulation, achieving a reliable estimate with 99.9\% confidence in just 52 iterations, compared to the \(24(n-1)\) iterations of the 2-point RANSAC used in~\cite{Ha2018-et}. 

\subsection{Step 2: Translation and Inverse Depth Estimation}\label{sec:step2}

We proceed by fixing the rotations \(\vec{\theta}_i^*\), estimated in section~\ref{sec:step1}, and initializing the translations as \(\vec{r}_i^{(0)} \leftarrow \bar{\vec{r}}_i^{*}/\Bar{w}\) and the inverse depths as \(w_j^{(0)} \leftarrow \Bar{w}\). 
These initial estimates are then refined by a BA procedure restricted to estimating only translations and inverse depths. 
We utilize the Georgia Tech Smoothing And Mapping Library (GTSAM)~\cite{gtsam} to implement and solve the restricted BA via factor graph optimization and a Levenberg Marquardt (LM) solver.
A custom factor is defined to enable simultaneous optimization of both \(\vec{r}_i\) and \(w_j\), encoding the image disparity residual.

To ensure all landmarks satisfy the cheirality condition, we reparameterize \(w_j\) using the soft-plus function $\mathtt{sp}(x) = {\ln\left(1+\exp\left(\alpha x\right) \right)}/{\alpha}$, such that $w_j=\mathtt{sp}(\omega_j)$, where $\omega_j\in \mathbb{R}$ is now the landmark-related optimization variable, initialized with $\omega^{(0)}_j \leftarrow \mathtt{sp}^{-1}(w^{(0)}_j)$.
To avoid numerical instability, by use the equation from~\cite{Wiemann2024-uy}, we have that 
\begin{equation}
    \mathtt{sp}(x) = \max \{0, x\} + \frac{\mathtt{log1p}(\exp{(-|\alpha x|)})}{\alpha},
    \label{eq:softplus}
\end{equation}
where $\mathtt{log1p}(x)$ is a numerically stable implementation of $\ln(1+x)$, particularly for $x\rightarrow0^{+}$.

We incorporate~Eq.~\eqref{eq:softplus} into a custom factor $\vec{\varepsilon}_{ij}^{\mathrm{step2}}$ that encodes, as residual, the 2D disparity between the measured image point $\vec{p}_{ij}^{\mathrm{m}}$ and the expected projection of the corresponding landmark. 
The residual is modeled as a function of the parameters $\vec{r}_i$ and $\omega_j$, while keeping $\vec{\theta}_i^*$ fixed, as in
\begin{align}
    \vec{\varepsilon}_{ij}^{\mathrm{step2}}(\vec{r_i}, &\,\omega_j) = \nonumber \\
    &\vec{p}_{ij}^{\mathrm{m}} - \left\langle \vec{K} \left(\left( I_3 + [\vec{\theta}_i^{*}]_\times \right) \vec{x}_{0j}^{\mathrm{m}} + \mathtt{sp}(\omega_j) \vec{r_i} \right) \right\rangle.
    \label{eq:factorStep2}
\end{align}

We apply~Eq.~\eqref{eq:factorStep2} to construct the restricted BA problem for poses $i=1,\ldots,n$ and landmarks $j=1,\ldots,m$, as in
\begin{equation}
    \min_{\{\vec{r}_i\}_{i=1}^{n}, \{\omega_j\}_{j=1}^{m}} \sum_{i=1}^{n} \sum_{j=1}^{m} \Omega( \|\vec{\varepsilon}_{ij}^{\mathrm{step2}}(\vec{r}_i, \omega_j)\|_{ \Sigma_{ij}}^{2} ),
    \label{eq:baStep2}
\end{equation}
where $\Omega(x)$ is the robust Huber cost, $\|\vec{\varepsilon}\|_{\Sigma} = \sqrt{ \vec{\varepsilon}^\top\Sigma^{-1}\vec{\varepsilon}}$, and $\Sigma_{ij}$ is the covariance of the measurement $j$ taken at pose $i$. We solve~Eq.~\eqref{eq:baStep2} using the LM iterative procedure.

The optimized \(\omega_j^*\) values are converted to \(w_j^*\) using~Eq.~\eqref{eq:softplus}, ensuring positive inverse depths.
The minimizers $\left(\{\vec{r}_i^*\}_{i=1}^{n}, \{w_j^*\}_{j=1}^{m}\right)$ resulting from~Eq.~\eqref{eq:baStep2} are used to initialize the procedure in Section~\ref{sec:step3}. The procedures from Sections~\ref{sec:step1} and~\ref{sec:step2} collectively avoid the solution ambiguity associated with pose estimation from essential matrix decomposition~\cite{nister2004efficient}.

\subsection{Step 3: Full Bundle Adjustment}\label{sec:step3}

Equipped with estimates for all parameters \(\vec{\theta}_i^{*}\), \(\vec{r}_i^{*}\), and \(w_j^{*}\), we solve a full BA to compute the camera trajectory and the coordinates of the landmarks. 

\begin{figure}[ht]
    \centering    
	\def\Elevation{32} 
	\def\Azimuth{88} 
	\includegraphics[width=3in]{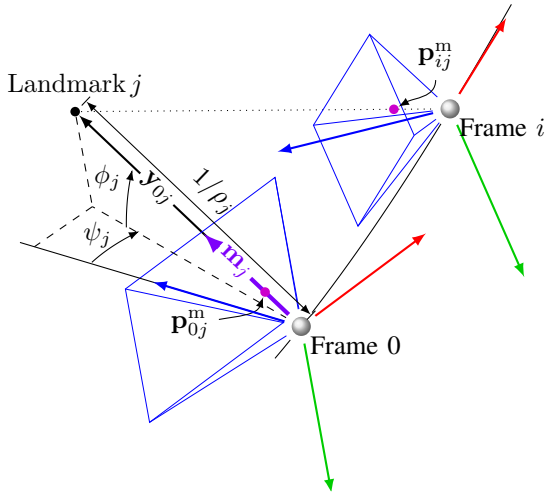}
    \caption{Landmark parameterization using inverse depth \(w\), azimuth \(\psi\), and elevation \(\phi\).}
    \label{fig:landmarks}
\end{figure}

Deviating from the landmark definition in Sections~\ref{sec:step1} and~\ref{sec:step2}, we reformulate the parameterization of the expected landmark coordinates to account for image feature quantization by introducing variables for in-camera-plane variation of the coordinates instead of directly exploiting the reference camera frame's image feature measurement.
We adopt the inverse depth approach in~\cite{Civera2008-xw}, Fig.~\ref{fig:landmarks}, re-parameterizing the 3D point $\vec{y}_{0j}$ as $\vec{y}_{0j} = \vec{m}_j/\rho_j$, where \(\vec{m}_j = \vec{m}(\psi_{j}, \phi_{j})\) is a unit directional vector from the reference camera frame to the \(j\)-th landmark.
Here, $\vec{m}(\psi,\phi)$ is defined as $\vec{m}(\psi,\phi)=[\cos{\phi}\sin{\psi}, -\sin{\phi}, \cos{\phi}\cos{\psi}]^\top$, and $\rho_j = 1/\|\vec{y}_{0j}\|$ is the inverse depth, replacing the prior definition $w_j=1/Z_{j}$.
Accordingly, \(\vec{m}_j\) is parameterized by the azimuth angle \(\psi_j\) and the elevation angle \(\phi_j\), which, for $\vec{y}_{0j} = [X_j,Y_j,Z_j]^\top$, are computed as $\psi_j = \arctan(X_{j}/ Z_{j})$ and $\phi_j = \arctan(-Y_{j}/ \sqrt{X_{j}^2 + Z_{j}^2})$.
As in Section~\ref{sec:step2}, we apply Eq.~\eqref{eq:softplus} in $\rho_j = \mathtt{sp}(\omega_j)$ to ensure $\rho_j$ remains strictly positive, guaranteeing the cheirality condition.

We define a custom factor $\vec{\varepsilon}_{ij}^{\mathrm{step3}}$ that encodes the residual described in Section~\ref{sec:step2}; however, the residual is now modeled as a function of the camera's rotation $\vec{R} \in \mathrm{SO}(3)$, the camera's translation $\vec{r} \in \mathbb{R}^3$, and the landmark parameters $\omega, \psi, \phi \in \mathbb{R}$, such that
\begin{equation}
\begin{aligned}
    \vec{\varepsilon}_{ij}^{\mathrm{step3}}(\vec{R}, \vec{r}, &\,\omega, \psi, \phi) = \\
    &\vec{p}_{ij}^{\mathrm{m}} - \left\langle \vec{K} \left( \vec{R} \vec{m}(\psi,\phi) + \mathtt{sp}(\omega) \vec{r} \right) \right\rangle.
\end{aligned}
\label{eq:factorStep3}
\end{equation}
Thus, $\rho_j$, $\psi_j$, and $\phi_j$ determine the 3D location of landmark $\vec{y}_{0j}$.

Meanwhile, we maintain the reference frame at a fixed pose with $\vec{R}_0 = I_3$ and $\vec{r}_0 = \vec{0}$.
Reformulating~Eq.~\eqref{eq:factorStep3}, for each reference frame measurement $\vec{x}_{0j}, \, j=1,\ldots,m$, we constrain $\psi_j$ and $\phi_j$ to yield an additional factor $\vec{\varepsilon}_{0j}^{\mathrm{prior}}$. The factor encodes the residual described in Section~\ref{sec:step2}, but now as a function of $\psi_j$ and $\phi_j$, as in
\begin{align}
    \vec{\varepsilon}_{0j}^{\mathrm{prior}} (\psi_j, \phi_j) = \vec{p}_{0j}^{\mathrm{m}} - \left\langle \vec{K} \vec{m}(\psi_j,\phi_j) \right\rangle.
    \label{eq:factorReFrameStep3}
\end{align}

We use~Eq.~\eqref{eq:factorStep3} and~Eq.~\eqref{eq:factorReFrameStep3} to construct the BA problem for poses $i=1,\ldots,n$ and landmarks $j=1,\ldots,m$, as in
\begin{equation}
    \begin{aligned}
        \min_{\substack{\{(\vec{R}_i,\vec{r}_i)\}_{i=1}^{n} \\ \{(\omega_j, \psi_j,\phi_j)\}_{j=1}^{m} }} 
        &\sum_{j=1}^{m} \Omega( \|\vec{\varepsilon}_{0j}^{\mathrm{prior}}(\psi_j, \phi_j)\|_{ \Sigma_{0j}}^{2}) \\
        + &\sum_{i=1}^{n} \Omega( \|\vec{\varepsilon}_{ij}^{\mathrm{step3}}(\vec{R}_i, \vec{r}_i, \omega_j, \psi_j, \phi_j)\|_{ \Sigma_{ij}}^{2}),
    \end{aligned}
    \label{eq:cost3}
\end{equation}
where $\Omega(x)$, $\|*\|_{\Sigma}$ and $\Sigma_{ij}$ are described in Section~\ref{sec:step2}.
The minimizers $\left(\{(\vec{R}_i^{**}, \vec{r}_i^{**})\}_{i=1}^{n}, \{(\omega_j^{**}, \psi_j^{**},\phi_j^{**})\}_{j=1}^{m}\right)$ to problem~Eq.~\eqref{eq:cost3} serve as the vSLAM initialization solution.

\section{Experiments}

We evaluate our initialization pipeline on 100 synthetic small-motion image sequences generated in a custom simulation platform in Unreal Engine 5~\cite{unrealengine}.
\begin{figure}
    \vspace{.5em}
    \centering
    \includegraphics[width=0.238\textwidth]{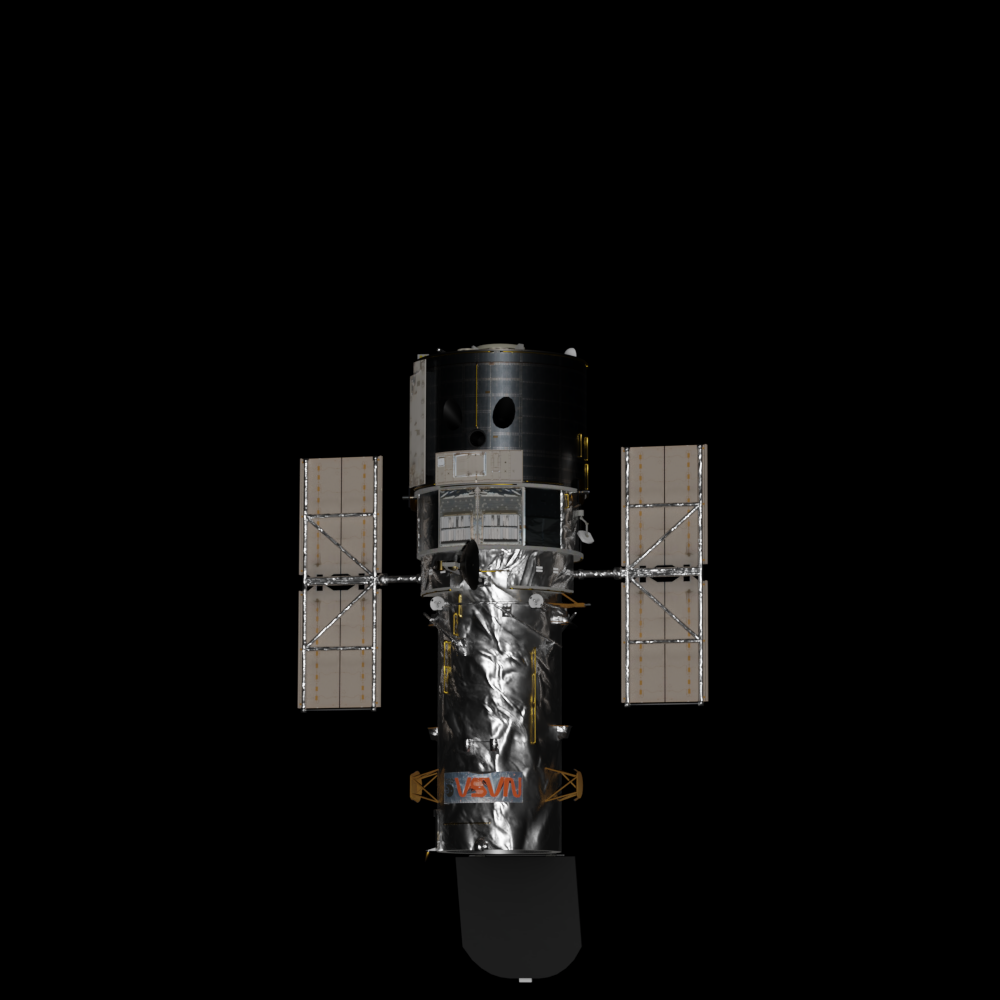}
    \hfill 
    \includegraphics[width=0.238\textwidth]{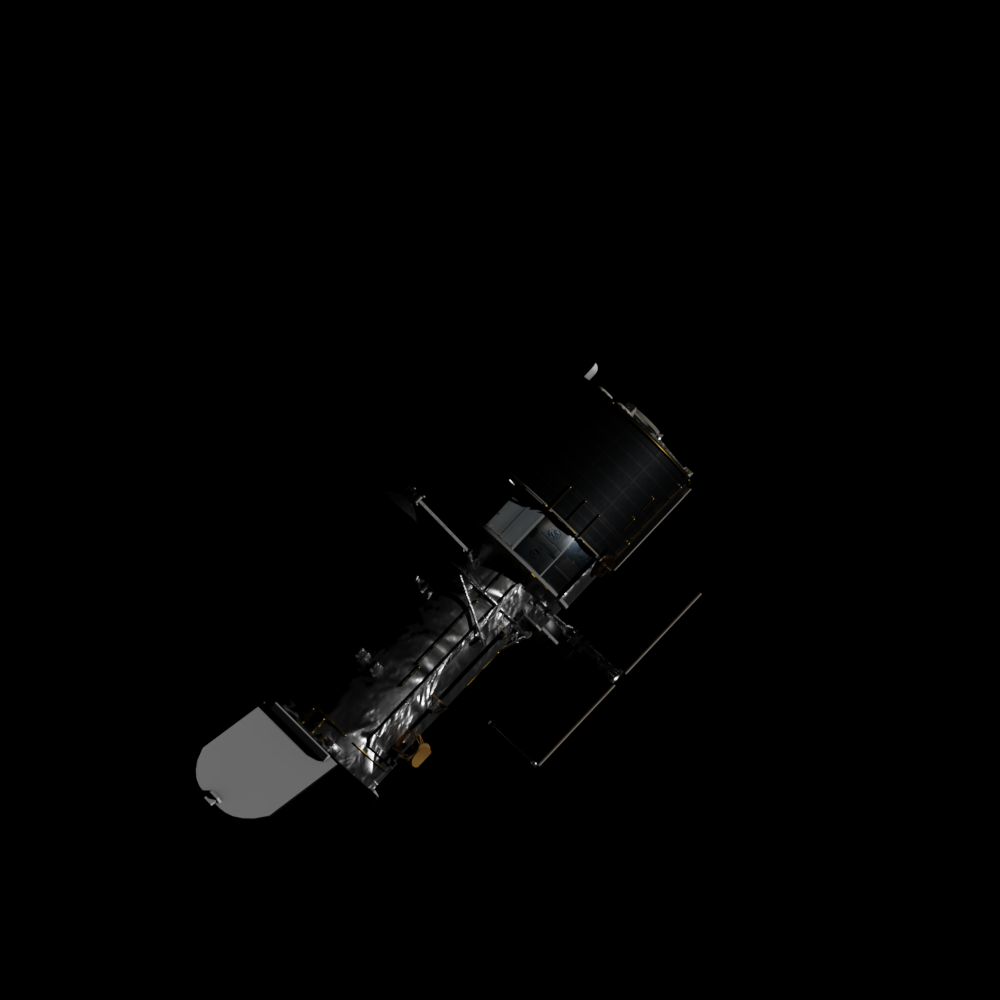}
    \caption{Realistic synthetic images of the Hubble Space Telescope produced in our Unreal Engine 5 simulator. As the camera and RSO move, the specular and diffuse reflections and moving shadows simulate the realistic, changing, and harsh illumination conditions encountered in space.}
    \vspace{-1.5em}
    \label{fig:simulated}
\end{figure}
Each sequence represents a weak-perspective, center-pointing RSO inspection trajectory and consists of 20 images captured at 10 frames per second using a simulated camera with a 14.9-degree field of view. 
The RSO is a tumbling, highly-detailed model of the Hubble Space Telescope (HST), positioned 100 meters from the camera. 
The simulated images of the tumbling HST, as shown in Fig.~\ref{fig:simulated}, exhibit dynamic illumination conditions that closely mimic the challenges of inspection in the vicinity of a non-cooperative RSO. 

\begin{figure}[ht]
    \vspace{.5em}
    \includegraphics[width=3.4in, clip=true, trim=0in 0.1in 0in 0.10in]{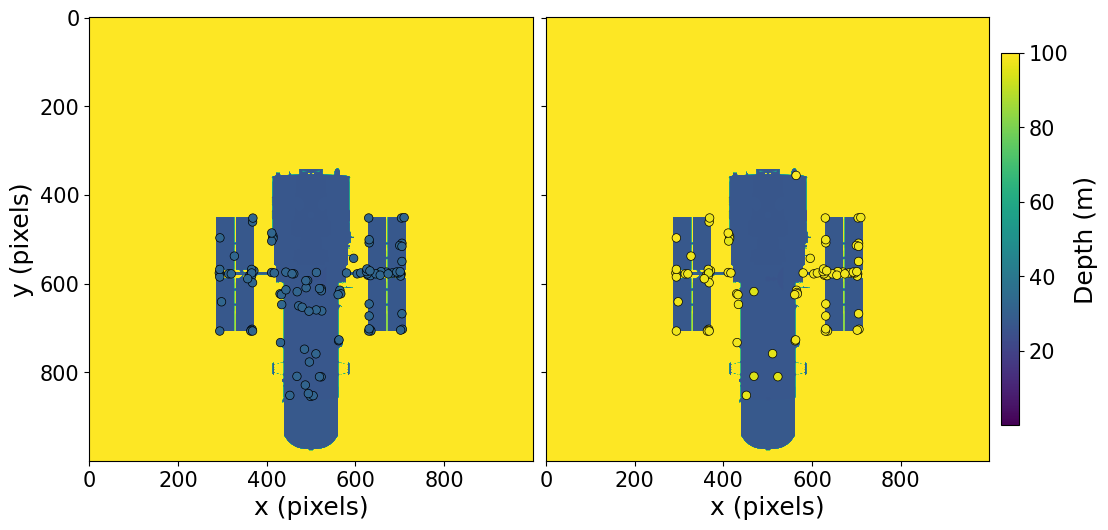}
    \begin{minipage}[t]{1.9in}
        \vspace{-1.5em}
        \raggedright
        \subcaption{Proposed}
    \end{minipage}
    \begin{minipage}[t]{0.8in}
        \vspace{-1.5em}
        \raggedleft
        \subcaption{Ha et al.}
    \end{minipage}
    \caption{The estimated landmark depth solutions superimposed and colored according to the depth maps of the HST model. The proposed method's depths are estimated closer to the ground truth than the method in~\cite{Ha2018-et}.}
    \vspace{-1.5em}
    \label{fig:depthmap}
\end{figure}

We compare the proposed method to the SfSM pipeline developed by Ha et al.~\cite{Ha2018-et} and the USAC\_FM\_8PTS algorithm~\cite{Raguram2013}.
For consistency, we align the reference frame of the estimated trajectory with that of the ground truth trajectory and we normalize the estimated translations by the magnitude of the $n$-th frame translation.
Depths are scaled by the same normalizing factor to ensure a consistent scale.
For each sequence, we compute (a) the absolute translation error $\vec{\epsilon}_{i}^{\mathrm{ATE}} = \vec{R}_i^{\top} \vec{r}_i -(\vec{R}_i^{\mathrm{true}})^{\top} \vec{r}_i^{\mathrm{true}}$ across all frames; (b) the absolute rotation error $\vec{\epsilon}_{i}^{\mathrm{ARE}} = \vec{\theta}_i - \vec{\theta}_i^{\mathrm{true}}$ across all frames; and (c) the depth error $\vec{\epsilon}_{j}^{\mathrm{depth}} = Z_j - Z_j^{\mathrm{true}}$ at the reference frame across all landmarks.
Depth error is obtained by comparing estimated depths against the scale normalized depth map of the simulated RSO, shown in Fig.~\ref{fig:depthmap}. We compute the root-mean-square (RMS) for each error value, given by
\begin{equation}
    \epsilon_{\mathrm{RMS}}^{*} = \sqrt{ \frac{\sum_{i=1}^{N} \|\vec{\epsilon}_{i}^{*}\|^2} {N} },
\end{equation}
where $N$ is the sample size and $* \in  \{\mathrm{ATE},\, \mathrm{ARE}, \, \mathrm{depth} \}$.
To represent data trends across all sequences, we compute the mean of the RMS errors.

\begin{table*}[t!]
 \vspace{1em}
 \caption{Comparison between the proposed method, ~\cite{Ha2018-et}, and~\cite{Raguram2013} for 100 weak-perspective, center-pointing, small-motion sequences.}
    \begin{tabular}{|>{\centering\arraybackslash}b{0.9in}|>{\centering\arraybackslash}b{0.8in}|>{\centering\arraybackslash}b{0.8in}|>{\centering\arraybackslash}b{0.8in}|>{\centering\arraybackslash}b{1.1in}|>{\centering\arraybackslash}b{1.5in}|}
        \hline
        \textbf{Method} & \textbf{Success Rate} & \textbf{RMS ATE} & \textbf{RMS ARE} & \textbf{Normalized Depth RMSE Mean} & \textbf{Mean Compute Time} \\
        (-) & (\%)& (m) & (deg) & (-) & (s)\\
        \hline
        Proposed & \textbf{62.0} & \textbf{0.277} & \textbf{0.032} & \textbf{0.605} & \textbf{0.040} - \textbf{0.156} - 55.8 \\
        \hline
        Ha et al.~\cite{Ha2018-et} & 39.0 & 0.818 & 0.112 &  1.25 & 0.107 - 0.192 - \textbf{21.2} \\
        \hline
        USAC~\cite{Raguram2013} & 0.0 & 0.970 & 0.046 & 4.11 & 0.035 \\
        \hline
    \end{tabular}
    \label{tab:pipeline_comparison}
    \vspace{-1.0em}
\end{table*}

All evaluated methods were implemented in C++ using the GTSAM library and OpenCV~\cite{opencv_library}.
We provided the pipelines with the same inlier ORB~\cite{rublee2011} feature tracks from a shared front-end. 
Furthermore, both SfSM methods used the identical LM optimizer parameters. 
The experiments were conducted on a machine equipped with an AMD Ryzen 7800X3D CPU and 32 GB of DDR5 RAM. 

The results, summarized in Table~\ref{tab:pipeline_comparison}, demonstrate that the proposed method is significantly more robust than both Ha's SfSM method and the state-of-the-art USAC\_FM\_8PTS algorithm when initializing from realistic inspection trajectories that emulate a tumbling RSO.

\begin{figure}[ht]
    \centering    
    \includegraphics[width=3.3in, clip=true, trim=0.1in 0.1in 0.1in 0.1in]{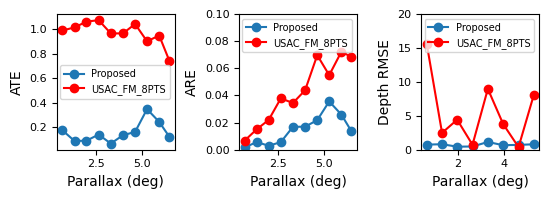}
    \caption{Error values for scale-normalized translations, rotations, and scale-normalized depths for the proposed method (blue) and the USAC\_FM\_8PTS algorithm~\cite{Raguram2013} (red) on trajectories spanning from 0 to 6 degrees of parallax.}
    \label{fig:usac}
\end{figure}
\begin{figure}[ht]
    \vspace{0.5em}
    \centering    
    \includegraphics[width=\linewidth]{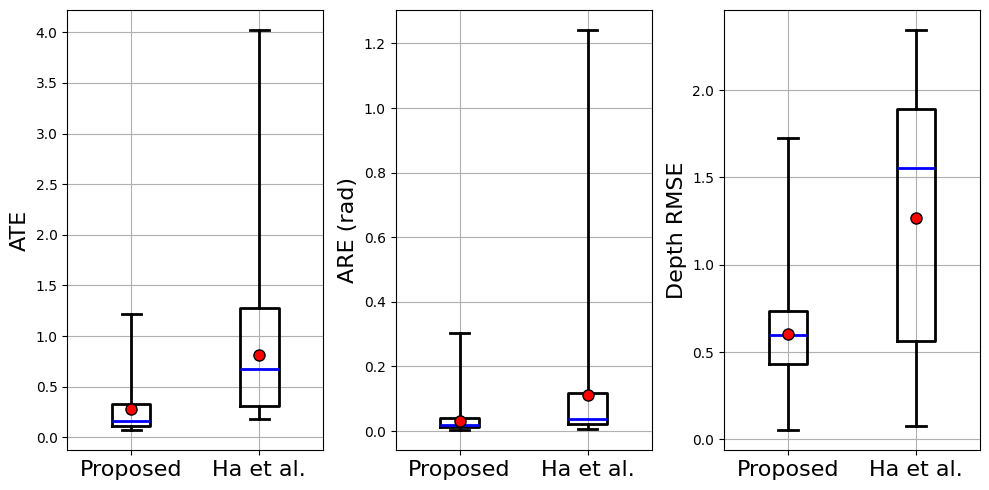}
    \caption{Error value distribution, median (blue line),  and mean (red circle) for scale-normalized translations (left), rotations (middle), and scale-normalized depths (right) for the proposed method and the method in~\cite{Ha2018-et}.}
    \label{fig:boxplot}
\end{figure}

\begin{figure}[ht]
    \begin{subfigure}[b]{1.92in}  
        \includegraphics[width=1.91in]{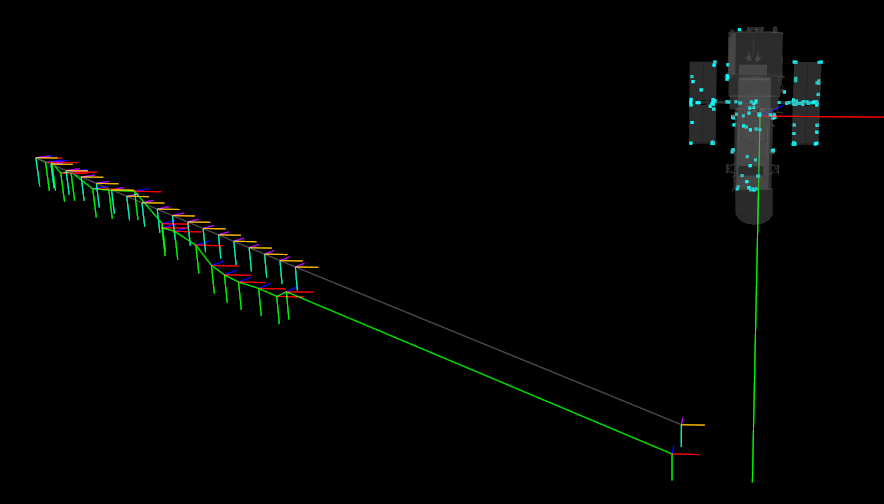}
        \caption{Pose trajectory}
        \label{fig:trajectory}
    \end{subfigure}
    \begin{subfigure}[b]{1.38in}  
        \includegraphics[width=1.37in]{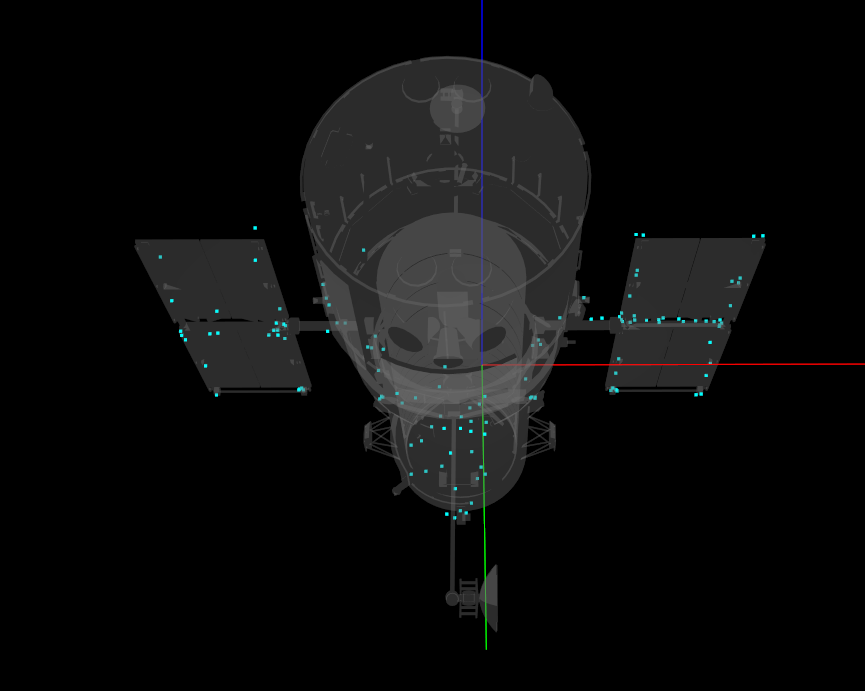}
        \caption{3D reconstruction}
        \label{fig:reconstruction}
    \end{subfigure}
    \caption{SLAM solution one step after initialization, with estimated poses (RGB) closely aligned with ground truth poses (CYM) and estimated landmarks (cyan) coinciding with the ground truth HST object.}
    \label{fig:nominal}
    \vspace{-2.2em} 
\end{figure}

The proposed method successfully initialized in 62\% of the sequences---a~$59$\% improvement over Ha's approach, which only initialized in 39\% of the sequences. 
In contrast, the USAC\_FM\_8PT method, tested at parallaxes up to 6 degrees, failed to initialize in any of the sequences and demonstrated lower accuracy than the proposed method, as shown in Fig.~\ref{fig:usac}.
In Fig.~\ref{fig:boxplot}, the smaller height of the box plots for our method indicates a lower spread in error across all SLAM variables, demonstrating more reliable and robust performance. 
Moreover, compared to Ha's approach, the proposed method is, on average, 66.1\% more accurate in translation estimation, 71.4\% more accurate in rotation estimation, and 51.6\% more accurate in depth estimation.
Thus, we are able to more effectively and consistently overcome the bas-relief ambiguity when initializing in RSO inspection trajectories, as seen in Fig.~\ref{fig:nominal}.

Although more robust and accurate, the proposed algorithm achieves these improvements at the cost of an increased step 3 computation time that is 1.63 times longer than the competing SfSM method.
The inverse depth and landmark re-parameterizations in step  3 enhance the robustness of the solution, but also introduce additional variables to the the optimization, increasing the dimensionality of the search space. 
A key challenge arises from the vanishing gradient issue, which hinders convergence~\cite{nocedal2006large}. 
The problem is exacerbated by ambiguous observations and scale differences between the optimization variables. 
However, when considering the scale and range of RSO inspection trajectories, the added computation time is less impactful.





\section{CONCLUSIONS}

In this work, we present a monocular initialization pipeline that extends the Ha et al. three-step SfSM method~\cite{Ha2018-et} to address visual estimation in RSO inspection trajectories, which are characterized by weak-perspective projection, center-pointing trajectories, dominant planar geometry, and dynamic illumination conditions.
We demonstrate improved accuracy and robustness over Ha's approach and the USAC\_FM\_8PTS wide-baseline method when evaluating on realistic simulated image sets.

Future work for the proposed method should focus on improving the computation time and further robustness of the initializer, enabling its integration into monocular SLAM pipelines for real-time, time-critical applications.
For instance, variable pre-conditioning may be used to increase the rate of convergence of step 3 of the pipeline, addressing the vanishing Jacobian issue.
Other potential improvements include using simulated annealing~\cite{kirkpatrick1983optimization} for better local minima exploration and applying convex relaxation methods~\cite{rosen2015convex} to improve nonlinear program initialization. 
Incorporating motion constraints for different RSO motion paradigms may also help further constrain the problem.






\bibliography{refs} 

\end{document}

%% file: preamble.tex
\usepackage{import}
\usepackage{graphicx}
\usepackage{amsmath}
\usepackage{amsfonts}
\usepackage{amssymb}
\usepackage{fmtcount}
\usepackage{mdframed}
\usepackage{transparent}
\usepackage{bm}
\usepackage{mathtools}
\usepackage{cancel}
\usepackage[export]{adjustbox}
\usepackage{tikz,pgfplots,pgfmath,tikzscale,wrapfig}
\usepackage{subcaption}
\usepackage[thinlines]{easytable}
\usepackage{tabularx}
\usepackage{placeins}
\usepackage{cite}

\usetikzlibrary{shapes.misc}
\usetikzlibrary{arrows, arrows.meta}
\usetikzlibrary{shapes.geometric,calc,positioning}
\usetikzlibrary{decorations.markings}
\pgfplotsset{compat=newest}
\usetikzlibrary{positioning}

\colorlet{bgcolor}{white}
\definecolor{dark-green}{rgb}{0.1328,0.5430,0.1328}

\let\vec\undefined

\newcommand{\harpoonvec}[2]{%
	\ifx\displaystyle#1\doalign{$\harpvecsign$}{#1#2}\fi
	\ifx\textstyle#1\doalign{$\harpvecsign$}{#1#2}\fi
	\ifx\scriptstyle#1\doalign{\scalebox{.6}[.9]{$\harpvecsign$}}{#1#2}\fi
	\ifx\scriptscriptstyle#1\doalign{\scalebox{.5}[.8]{$\harpvecsign$}}{#1#2}\fi
}
\newcommand{\harpvecsign}{\scriptscriptstyle\rightharpoonup}

\newcommand{\doalign}[2]{%
	{\vbox{\offinterlineskip\ialign{\hfil##\hfil\cr#1\cr$#2$\cr}}}%
}

\newcommand{\vec}[1]{\bm{\mathrm{#1}}}

\tikzset{
	overdraw/.style={preaction={draw,bgcolor,line width=#1}},
	overdraw/.default=3pt
}

\definecolor{black}{RGB}{0,0,0}
\definecolor{light-gray}{gray}{0.95}
\definecolor{goldenrod}{RGB}{254,236,167}
\definecolor{cadetblue}{RGB}{137,136,152}
\definecolor{navyblue}{RGB}{25,114,180}
\definecolor{forestgreen}{RGB}{0,102,0}
\definecolor{pinegreen}{RGB}{18,138,114}
\definecolor{springgreen}{RGB}{198,218,110}
\definecolor{maroon}{RGB}{153,0,0}
\definecolor{brown}{RGB}{153,76,0}
\definecolor{orange}{RGB}{255,128,0}
\definecolor{denim}{RGB}{0,102,204}
\definecolor{darkblue}{RGB}{0,0,102}
\definecolor{blue}{RGB}{0,0,255}
\definecolor{indigo}{RGB}{125,0,250}
\definecolor{purple}{RGB}{204,0,204}
\definecolor{dark-yellow}{RGB}{102,102,0}
\definecolor{dark-teal}{RGB}{0,153,153}
\definecolor{green}{RGB}{0,204,0}

\tikzset{
  dim above/.style={to path={\pgfextra{
        \pgfinterruptpath
        \draw[>=latex,|<->|] let
        \p1=($(\tikztostart)!2mm!90:(\tikztotarget)$),
        \p2=($(\tikztotarget)!2mm!-90:(\tikztostart)$)
        in(\p1) -- (\p2) node[pos=.5,sloped,above]{#1};
        \endpgfinterruptpath
      }(\tikztostart) -- (\tikztotarget) \tikztonodes
    }
  },
  dim below/.style={to path={\pgfextra{
        \pgfinterruptpath
        \draw[>=latex,|<->|] let 
        \p1=($(\tikztostart)!2mm!90:(\tikztotarget)$),
        \p2=($(\tikztotarget)!2mm!-90:(\tikztostart)$)
        in (\p1) -- (\p2) node[pos=.5,sloped,below]{#1};
        \endpgfinterruptpath
      }(\tikztostart) -- (\tikztotarget) \tikztonodes
    }
  },
}

\makeatletter
\long\def\@makecaption#1#2{%
\ifx\@captype\@IEEEtablestring%
    \begin{center}{\footnotesize #1}\\{\footnotesize\scshape #2}\end{center}%
    \@IEEEtablecaptionsepspace
\else
    \@IEEEfigurecaptionsepspace
    \setbox\@tempboxa\hbox{\footnotesize #1.~#2}%
    \ifdim \wd\@tempboxa >\hsize%
        \setbox\@tempboxa\hbox{\footnotesize #1.~}%
        \parbox[t]{\hsize}{\footnotesize \noindent\unhbox\@tempboxa#2}%
    \else%
        \ifcenterfigcaptions \hbox to\hsize{\footnotesize\hfil\box\@tempboxa\hfil}%
        \else \hbox to\hsize{\footnotesize\box\@tempboxa\hfil}%
        \fi
    \fi
\fi}
\makeatother